\documentclass{article}
\usepackage{spconf,amsmath,graphicx}
\usepackage{cite}
\usepackage{siunitx}
\usepackage{amsmath,amssymb,amsfonts}
\usepackage{textcomp}
\usepackage{booktabs, makecell, multirow, tabularx}
\usepackage{cite}
\usepackage{url}
\usepackage{amsmath,amssymb,amsfonts}
\usepackage{algorithm}
\usepackage{graphicx}
\usepackage{changepage}
\usepackage{algpseudocode}
\usepackage{gensymb}
\usepackage[hidelinks]{hyperref} 
\usepackage{lipsum}
\newcommand\mytab[1]{\begin{tabular}[t]{@{}c@{}} #1 \end{tabular}}
\newcommand\mc[2]{\multicolumn{#1}{c}{#2}}

\usepackage{multicol}
\usepackage{subcaption}
\title{FOOD: Facial Authentication and Out-of-distribution detection with short-range FMCW Radar}
\name{Sabri Mustafa Kahya$^{\star}$  \qquad Boran Hamdi Sivrikaya$^{\star}$ \qquad Muhammet Sami Yavuz$^{\star \dagger}$ \qquad Eckehard Steinbach$^{\star}$}

\address{  $^{\star}$Technical University of Munich,
School of Computation, Information and Technology,\\  Department of Computer Engineering, Chair of Media Technology,\\ Munich Institute of Robotics and Machine Intelligence (MIRMI) \\ $^{\dagger}$Infineon Technologies AG}

\begin{document}
%\ninept

\maketitle
\begin{abstract}
This paper proposes a short-range FMCW radar-based facial authentication and out-of-distribution (OOD) detection framework. Our pipeline jointly estimates the correct classes for the in-distribution (ID) samples and detects the OOD samples to prevent their inaccurate prediction. Our reconstruction-based architecture consists of a main convolutional block with one encoder and multi-decoder configuration, and intermediate linear encoder-decoder parts. Together, these elements form an accurate human face classifier and a robust OOD detector. For our dataset, gathered using a $\SI{60}{\giga\hertz}$ short-range FMCW radar, our network achieves an average classification accuracy of 98.07\% in identifying in-distribution human faces. As an OOD detector, it achieves an average Area Under the Receiver Operating Characteristic  (AUROC) curve of 98.50\% and an average False Positive Rate at 95\% True Positive Rate (FPR95) of 6.20\%. Also, our extensive experiments show that the proposed approach outperforms previous OOD detectors in terms of common OOD detection metrics.
\end{abstract}


\section{Introduction}
\label{sec:intro}

In recent years, short-range radars have attracted significant attention and widespread adoption across various domains due to their cost-effectiveness. Radar systems demonstrate remarkable resilience to adverse environmental conditions, including lightning, rain, and smoke. Moreover, they are not constrained by privacy issues, which provides a distinct advantage over camera-based applications. Therefore, they find extensive use in indoor applications such as human presence detection, human activity classification, people counting, gesture recognition, and even heartbeat estimation \cite{human_presence7,human_activity3,people_counting,gesture_recog,heartbeat_est}. In this study, we propose a facial authentication system, FOOD, which also addresses the detection of out-of-distribution (OOD) samples.

OOD detection plays a critical role in ensuring the secure and reliable deployment of deep learning (DL) models in real-world scenarios by preventing overconfident decisions on samples that deviate from the training data 
 \cite{b9,b5,b27,b25}. FOOD provides a parallel solution for both classification and OOD detection tasks.

FOOD is a novel facial authentication system that not only classifies in-distribution (ID) human face samples but also detects OOD face samples that were not presented during training. The system utilizes a unique framework that processes raw ADC radar data to accurately classify three human faces and detect OOD samples. Our reconstruction-based framework provides convolutional and linear encoder-decoder parts (see Figure \ref{fig:pipline}). We call the linear encoder-decoder parts the leaves of the network. Namely \textbf{common leaf (CL)} and \textbf{private leaves (PLs)}. They are mainly designed for OOD detection purposes, but the \textbf{PLs} also strongly affect accurate classification together with the main convolutional part. A standard OOD detector utilizes a scoring function along with a predefined threshold. Subsequently, a test sample is categorized as ID if its score falls below the specified threshold; otherwise, it is classified as OOD. In this study, we use multi-thresholding. Here are the key contributions of this study: 

\begin{itemize}
\item We introduce a novel reconstruction-based architecture designed to function as both a robust classifier and an OOD detector. It comprises a main convolutional one-encoder multi-decoder segment denoted as \textbf{MP} and intermediate linear one-encoder one-decoder components, namely \textbf{CL} and \textbf{PLs}. Within \textbf{MP}, the multi-decoder section is dedicated to identifying specific human face classes: Person 1 (\textbf{PER$_1$}), Person 2 (\textbf{PER$_2$}), and Person 3 (\textbf{PER$_3$}). \textbf{CL} is situated at the termination of the \textbf{MP}'s encoder, while \textbf{PLs} are strategically positioned within the decoders of \textbf{MP}. \textbf{CL} and \textbf{PLs} are mainly designed for OOD human face sample detection, but \textbf{PLs} also significantly influence overall classification accuracy.

\item We propose a novel loss function composed of seven distinct reconstruction losses. Among these, three originate from the \textbf{MP},  while one is derived from the \textbf{CL}. The remaining three losses are associated with the \textbf{PLs}. During test time, the \textbf{MP} and \textbf{PL} losses are utilized for classification purposes, while the \textbf{CL} and \textbf{PL} losses are employed for OOD detection purposes.

\item We conduct comprehensive experiments that yield an average human facial classification accuracy of 98.07\% and an average AUROC of 98.50\% in OOD detection. Additionally, we compare FOOD's performance with state-of-the-art (SOTA) OOD detection methods, evaluating it using common OOD detection metrics and test time, and demonstrate its superior performance.

\item We also perform extensive ablation studies to emphasize the effects of the novel parts of our network. We show how \textbf{PLs} affect the classification accuracy when they are used together with the main convolutional reconstruction part. Also, we demonstrate the results of OOD detection when we only utilize \textbf{CL} and \textbf{CL} together with \textbf{PLs}.

\end{itemize}

\section{Related Work}
\label{sec:related}

In the field of radar-based face authentication, various studies have been conducted. The work in \cite{dnn-based} used a $\SI{61}{\giga\hertz}$ mmWave radar sensor, employing a deep neural network (DNN) to classify human faces. The authors of \cite{dnn-based} captured signals from multiple antenna elements, with data gathered from eight individuals at varying distances and angles relative to the radar. By combining signals from each receiver, a classification accuracy of 92\% was achieved. Similarly, \cite{cnn-based} introduced a $\SI{61}{\giga\hertz}$ FMCW radar face identification system using a Convolutional Neural Networks (CNNs). Data for this system were collected from three individuals positioned 30 cm from the radar chip, both with and without cotton masks. Initially, the model was trained on subjects not wearing masks, and subsequently, data from subjects wearing cotton masks were added. The study reports findings for both scenarios. Another significant contribution is from \cite{32by32}, which employs a radar sensor with 32 transmit (Tx) and 32 receive (Rx) antennas. A dense auto-encoder model was utilized on a dataset comprising 200 distinct faces, training this system as a one-class classifier for a verification system tailored to each person. Following this, the authors in \cite{imp32by32} also used the same dataset but proposed an improvement by integrating a convolutional autoencoder with a random forest classifier. \cite{one-shot} proposed a novel approach by introducing a one-shot learning method for a $\SI{61}{\giga\hertz}$ FMCW radar. Based on a Siamese Network architecture, this method utilized data from eight individuals positioned at distances ranging from 30 to 50 cm from the radar, achieving an accuracy of 97.6\%. The most recent study, \cite{point-cloud-face}, proposed a face recognition system using a $\SI{77}{\giga\hertz}$ mmWave sensor. This system extracts a point cloud dataset from nine users at various distances (60 cm and 80 cm) and angles (-45\degree, 0\degree, 45\degree). The architecture is based on PointNet and adapted to work directly with the sparser point cloud data from mmWave radar sensors. The device used in this study is a cascade device, comprising four radar chips, each with 3 Tx and 4 Rx antennas, resulting in a total of 12 Tx and 16 Rx antennas, and achieving an accuracy of 98.69\%. The studies mentioned, however, predominantly concentrate on achieving a highly accurate classifier, overlooking the performance of their pipeline when exposed to OOD samples.

OOD detection was first introduced in \cite{b1}, which employed maximum softmax probabilities to differentiate OOD input from ID samples by claiming that OOD instances typically exhibit lower softmax scores than ID samples. ODIN \cite{b2} aimed to enhance ID softmax scores through input perturbation and temperature scaling of logits. A model ensembling technique was layered on ODIN's approach in \cite{b3}. G-ODIN \cite{b30} extended ODIN's methodology \cite{b2} with an innovative training strategy. Meanwhile, MAHA \cite{b4} introduced a Mahalanobis distance-based OOD detection using intermediate layer representations. Similarly, FSSD \cite{b6} leverages these intermediate representations for detection. The approach in \cite{b31} employs a non-parametric KNN on penultimate layer embeddings for detection. An energy-based OOD detection method was introduced by \cite{b7}, utilizing the $logsumexp$ function. ReAct \cite{b28} implements truncation on activations in the penultimate layer and it is compatible with various OOD detection methods. GradNorm \cite{b14} distinguishes between IDs and OODs using the vector norm of gradients backpropagated from the KL divergence between the softmax output and a uniform distribution. In \cite{hendrycks2022scaling}, two methods were proposed: MaxLogit, which distinguishes between ID samples and OOD samples based on maximum logit scores, and KL, which utilizes minimum KL divergence information. The OE technique in \cite{b8} employs limited OOD examples during training to align their softmax scores closer to a uniform distribution. OECC \cite{b10} enhances OE with a novel loss incorporating extra regularization elements. 

OOD detection has also been explored in the radar domain. \cite{RB-OOD} employs $\SI{60}{\giga\hertz}$ FMCW radar to detect commonly seen indoor objects as OOD rather than a walking person. MCROOD \cite{MCROOD} serves as a multi-class radar OOD detector that distinguishes moving OOD objects from human activities of sitting, standing, and walking. \cite{kahya2023hood} introduced a multi-encoder multi-decoder network, which handles both human presence and OOD detection concurrently. \cite{kahya2023harood} also focuses on human activity classification and OOD detection.

\section{Radar System Design}

In this research, we used Infineon's BGT60TR13C $\SI{60}{\giga\hertz}$ FMCW radar chipset for data acquisition. The radar configuration details are provided in Table \ref{tab:radar_conf}. This chipset has a single Tx antenna and three Rx antennas. During the transmission, the Tx antenna emits \(N_c\) chirp signals. These signals, upon interaction with objects in the environment, are reflected and subsequently received by the Rx antennas with a time delay, indicative of the object’s range and velocity. The transmitted and received signals are then mixed and passed through a low-pass filter to extract the intermediate frequency (IF) signal. Subsequently, this IF signal is converted into digital form using an Analog-to-Digital Converter (ADC) that operates at a sampling frequency of $\SI{2}{\mega\hertz}$ with $12$-bit accuracy. The raw ADC data is structured as \(N_{Rx} \times N_c \times N_s\) and used as input for our model.

\iffalse
During the transmission, the Tx antenna emits \(N_c\) chirp signals which is performed by the calibration of a voltage-controlled oscillator (VCO), which operates in tandem with a phase-locked loop (PLL) set to an $\SI{80}{\mega\hertz}$ reference frequency. The chirp signal's mathematical representation is given by
\[
s(t) = \exp\left(j2\pi \left(f_c t + \frac{S}{2} t^2\right)\right), \quad \forall 0 < t < T_c
\]
where \(f_c\) denotes the center frequency and \(S\) is the chirp rate, defined as the ratio of the signal bandwidth \(B\) to the chirp duration (\(T_c\)). 

The chipset can generate highly linear frequency chirps within the $\SI{57}{\giga\hertz}$ to $\SI{64}{\giga\hertz}$ range. The calibration process involves adjusting the divider value and a tuning voltage between $\SI{1}{\volt}$ to $\SI{4.5}{\volt}$. These signals, upon interaction with environmental objects, are reflected and subsequently received by the Rx antennas with a time delay, indicative of the object’s range and velocity. The transmitted and received signals are then mixed and passed through a low-pass filter to extract the intermediate frequency (IF) signal. Subsequently, this IF signal is converted into digital form using an Analog-to-Digital Converter (ADC) that operates at a sampling frequency of $\SI{2}{\mega\hertz}$ with $12$-bit accuracy. The raw ADC data, which forms the core of our analysis, is structured as \(N_{Rx} \times N_c \times N_s\) and used as input for our model.
\fi

\begin{table}[h]
    \caption{\small FMCW Radar Configuration Parameters }
    \centering
     \footnotesize  
    \begin{tabular}{@ {\extracolsep{10pt}} ccc}
    \toprule

    \centering
    Configuration name & Symbol & Value \\
    \midrule
    Number of transmit antennas & $N\textsubscript{Tx}$  & 1  \\
    Number of receive antennas & $N\textsubscript{Rx}$  & 3  \\
    Chirps per frame & $N\textsubscript{c}$ & 64  \\
    Samples per chirp & $N\textsubscript{s}$  & 128 \\
    Frame period & T\textsubscript{f} & 50 \si{\ms}  \\
    Chirp to chirp time & $T\textsubscript{cc}$ & 391.55 \si{\us} \\
    Ramp start frequency & $f\textsubscript{min}$ & $\SI{60.1}{\giga\hertz}$\\ Ramp stop frequency & $f\textsubscript{max}$ &  $\SI{61.1}{\giga\hertz}$\\
    Bandwidth & $B$ & $\SI{1}{\giga\hertz}$\\ 
    \bottomrule
    \end{tabular}

    \label{tab:radar_conf}
\end{table}

\vspace{-0.4cm}
\section{Problem Statement and FOOD}
\begin{figure*}[htbp]
\centerline{\includegraphics[width=0.72\linewidth]{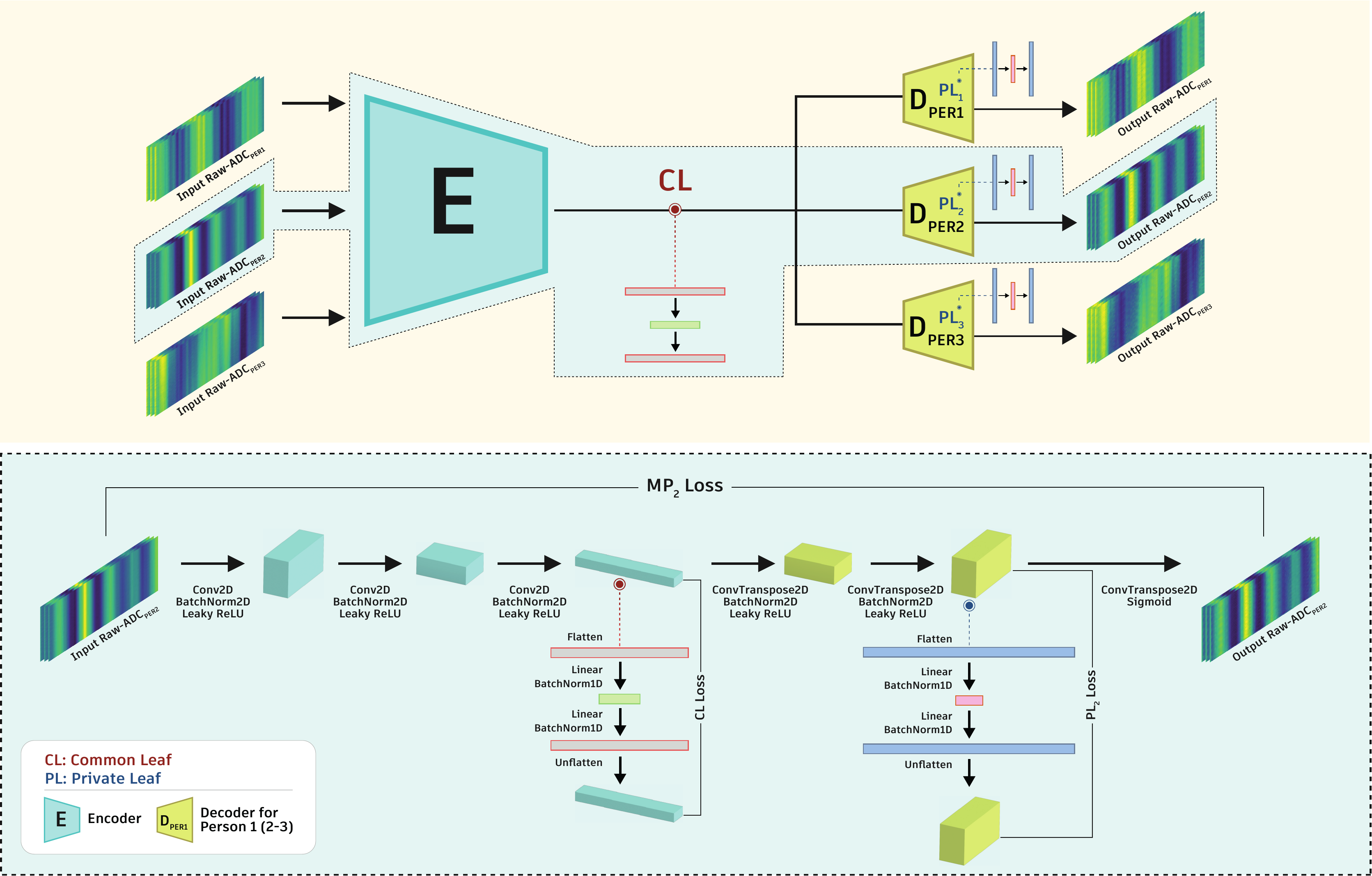}}
\caption{ \small The figure presents the high-level structure of FOOD and the zoom-in version of the highlighted section. Here we have a main \textbf{convolutional one-encoder multi-decoder part (MP)} and intermediate linear encoder decoder parts \textbf{common leaf (CL)} and \textbf{private leaves (PLs)}. The encoder of \textbf{MP} encodes the ID input data and from top to bottom its decoders reconstruct the ID input \textbf{PER$_1$}, \textbf{PER$_2$}, and \textbf{PER$_3$}, respectively. \textbf{CL} has a simple linear encoder-decoder network and is responsible for OOD detection. \textbf{PLs} are ID class specific and also have linear encoder-decoder network each. They have a considerable effect on both classification and OOD detection. MSE function is used for the calculation of each loss. }
\label{fig:pipline}
\end{figure*}

In this work, we address both multi-class classification and binary OOD detection tasks. FOOD provides a solution for accurately classifying ID face samples while correctly eliminating human faces (OODs) that were not present during training.

The \textbf{MP} consists of one encoder and three decoders. The encoder has three input gates and is responsible for encoding the three type ID samples. On the other hand, the top, middle, and bottom decoders are responsible for the reconstruction of the ID samples \textbf{PER$_1$}, \textbf{PER$_2$}, and \textbf{PER$_3$}, respectively. We calculate three reconstruction losses from \textbf{MP$_1$},\textbf{MP$_2$}, and \textbf{MP$_3$}, mainly used to achieve better classification performance. For example, \textbf{MP$_2$} refers to the encoder ($E$) and decoder of PER2 (D\textsubscript{PER2}) in \textbf{MP}, which is also detailed in the bottom part of Figure \ref{fig:pipline}.  Since we use the mean-squared-error (MSE) function, the three losses from this part are shaped as in the equation below:

\begin{equation*}
\begin{aligned}
\mathcal{L}_{MP}  = \frac{1}{b} \sum_{j \in \{p_1,p_2,p_3\}} \sum_{i=1}^{b} (\textbf{X}^{(i)}_{j} - D_{j}(E(\textbf{X}^{(i)}_{j})))^2,
    \end{aligned}
\label{eq:loss_recons}
\
\end{equation*}
where ${X}^{(i)}_{j}$ is raw ADC input, $b$ is the batch size, and $j$ is an index for the ID classes (\textbf{PER$_1$}, \textbf{PER$_2$}, \textbf{PER$_3$}); $E$ and $D_j$ correspond to the encoder and decoders of \textbf{MP}.

The \textbf{CL} starts at the end of the encoder of the \textbf{MP}. It has a simple linear encoder-decoder architecture and is responsible for encoding and reconstructing the intermediate features of all ID classes. The \textbf{CL} calculates a reconstruction MSEs to be used for better OOD detection:

\begin{equation*}
\begin{aligned}
\mathcal{L}_{CL}  = \frac{1}{b} \sum_{j \in \{p_1,p_2,p_3\}} \sum_{i=1}^{b} (E(\textbf{X}^{(i)}_{j}) - D_{CL}(E_{CL}(E(\textbf{X}^{(i)}_{j}))))^2,
\end{aligned}
\label{eq:loss_recons}
\
\end{equation*}
where $E_{CL}$ and $D_{CL}$ are the encoder and decoder of common leaf, respectively.

The \textbf{PLs} start just before the final reconstruction layer of each decoder. Since we have three decoders in \textbf{MP}, there are three \textbf{PLs}. Each has a simple linear encoder-decoder architecture and is responsible for encoding and reconstructing the intermediate features of their corresponding ID classes. The \textbf{PLs} calculate three reconstruction MSEs to be used for both more accurate classification and stronger OOD detection:

\begin{equation*}
\begin{aligned}
\mathcal{L}_{PL}  = \frac{1}{b} \sum_{k \in \{pl_1,pl_2,pl_3\}} \sum_{i=1}^{b} (\textbf{I}^{(i)}_{k} - D_{k}(E_{k}(\textbf{I}^{(i)}_{k})))^2,
\end{aligned}
\label{eq:loss_recons}
\
\end{equation*}
where ${I}^{(i)}_{k}$ represents the intermediate input of \textbf{PLs}, $k$ is an index for private leaves; $E_k$ and $D_k$ respectively correspond to the encoder and decoder of each \textbf{PL}.

\textbf{The final loss function} becomes $\mathcal{L}  =\mathcal{L}_{MP} + \mathcal{L}_{CL} + \mathcal{L}_{PL}$. We perform simultaneous training by minimizing all seven reconstruction losses at the same time. We use the MSE function to calculate the main and intermediate reconstruction losses by incorporating the Adamax optimizer \cite{kingma2014adam}.

\subsection{OOD Detection}
We use \textbf{CL} and \textbf{PLs} for OOD detection purposes. As a typical OOD detection task, we train our network only with IDs. Therefore, we expect less reconstruction errors coming from \textbf{CL} and \textbf{PLs} for ID samples than for OOD samples. We define three thresholds (guaranteeing 95\% of ID data is correctly detected) belonging to each ID class using \textbf{CL$+$PL$_1$} for \textbf{PER$_1$}, \textbf{CL$+$PL$_2$} for \textbf{PER$_2$}, and \textbf{CL$+$PL$_3$} for \textbf{PER$_3$}. During test time, a test sample passes through the entire network and gets three scores to be compared with the pre-defined thresholds explained above. The sample is classified as OOD if all three reconstruction scores from \textbf{CL}$+$\textbf{PLs} exceed their corresponding thresholds.

\subsection{Human Face Classification}
If a test sample is not classified as OOD, it is an ID. To define the correct class of the ID sample, we use \textbf{PLs} and \textbf{MP} (MP$_1$,MP$_2$, and MP$_3$). We separately calculate three error values from \textbf{MP$_1$$+$PL$_1$}, \textbf{MP$_2$$+$PL$_2$}, and \textbf{MP$_3$$+$PL$_3$}. We select the smallest one from these values and assign the ID sample to its corresponding class. For instance, if the total reconstruction MSE from \textbf{MP$_2$$+$PL$_2$} is less than the other two, the ID sample is classified as \textbf{PER$_2$}.

\begin{table*}[ht]
\centering
\small
\caption{\small OOD Detection Results. Performance comparison with SOTA methods. All values are shown in percentages. $\uparrow$ indicates that higher values are better, while $\downarrow$ indicates that lower values are better.}
\label{ood-results}
\setlength\tabcolsep{0pt}
\begin{tabular*}{\textwidth}{@{\extracolsep{\fill}}c|ccccccccccccc}
    \toprule
     
    & \mc{4}{\mytab{PER$_1$}} & \mc{4}{\mytab{PER$_2$}}
    & \mc{4}{\mytab{PER$_3$}} &{Test Time }\\
  
    \cmidrule{2-5} \cmidrule{6-9} \cmidrule{10-13}
\centering
     Methods 
    & AUROC
    & AUPR\textsubscript{IN}
    & AUPR\textsubscript{OUT}
    & FPR95 
    & AUROC 
    & AUPR\textsubscript{IN}
    & AUPR\textsubscript{OUT}
    & FPR95 
    & AUROC 
    & AUPR\textsubscript{IN}
    & AUPR\textsubscript{OUT}
    & FPR95 
    & (seconds) \\
    &  $\uparrow$
    &  $\uparrow$
    &  $\uparrow$
    &  $\downarrow$
    &  $\uparrow$
    &  $\uparrow$
    &  $\uparrow$
    &  $\downarrow$
    &  $\uparrow$
    &  $\uparrow$
    &  $\uparrow$
    &  $\downarrow$
    &  $\downarrow$ \\  
      
    \midrule
    MSP \cite{b1} &53.12&30.89&76.16&84.14&54.98&30.65&78.80&81.55&75.64&59.95&88.17&75.49&64  \\
    
    ODIN \cite{b2}  &52.56&30.29&75.96&84.26&53.76&29.90&78.40&81.65&77.53&62.92&88.98&75.02&255\\

     ENERGY \cite{b7} &39.00& 23.45&67.11&92.73&52.68&29.70&76.33&86.46&72.56&52.78&86.61&78.35&64  \\

    OE\cite{b8} & 53.43&29.83&72.05&95.58&56.67&29.41&79.47&82.91&69.63&39.91&86.36&70.76&63 \\
    REACT\cite{b28} &50.55&  31.00&  68.02&99.15&48.76&27.02&72.97&90.73&56.81&29.75&78.69&86.44&65\\
    GRADNORM\cite{b14} &77.77&61.03&88.46&70.45 &90.25 &76.51 &95.78&34.01&\textbf{100}&\textbf{98.95}&99.60&\textbf{0}&191 \\        
    MAXLOGIT\cite{hendrycks2022scaling} &48.58&28.51&70.92&92.32&51.82&28.97&74.77&91.04&60.01&33.10&81.46&82.10&64  \\
    KL\cite{hendrycks2022scaling}&52.75&31.67&72.10&92.57
    &57.28&32.49&77.90&88.20&67.99&41.06&85.25&76.37&63  \\
    \midrule
    \midrule
    \addlinespace

FOOD&\textbf{98.40}&\textbf{97.08}&\textbf{99.24}&\textbf{8.93}&\textbf{97.90}&\textbf{97.31}&\textbf{97.72}&\textbf{6.14}&99.25&98.45&\textbf{99.67}&3.20&\textbf{5} \\

    %   \rowcolor{lightgray}
    
    \bottomrule

\end{tabular*}
\end{table*}

\begin{table*}[ht]
\centering
\small
\caption{\small Ablation study for OOD detection. All values are shown in percentages. $\uparrow$ indicates that higher values are better, while $\downarrow$ indicates that lower values are better.}
\label{ablation}
\setlength\tabcolsep{0pt}
\begin{tabular*}{\textwidth}{@{\extracolsep{\fill}}c|ccccccccccccc}
    \toprule
     
    & \mc{4}{\mytab{PER$_1$}} & \mc{4}{\mytab{PER$_2$}}
    & \mc{4}{\mytab{PER$_3$}} &{}\\
  
    \cmidrule{2-5} \cmidrule{6-9} \cmidrule{10-13}
\centering
      
    & AUROC
    & AUPR\textsubscript{IN}
    & AUPR\textsubscript{OUT}
    & FPR95 
    & AUROC 
    & AUPR\textsubscript{IN}
    & AUPR\textsubscript{OUT}
    & FPR95 
    & AUROC 
    & AUPR\textsubscript{IN}
    & AUPR\textsubscript{OUT}
    & FPR95 
    &  \\
    &  $\uparrow$
    &  $\uparrow$
    &  $\uparrow$
    &  $\downarrow$
    &  $\uparrow$
    &  $\uparrow$
    &  $\uparrow$
    &  $\downarrow$
    &  $\uparrow$
    &  $\uparrow$
    &  $\uparrow$
    &  $\downarrow$ 
    &   \\  
      
    \midrule
 CL&94.21&90.15&97.09&32.47&93.25&89.75&94.75&37.16&94.99&91.22&97.52&29.24&\\     
CL + PLs&\textbf{98.40}&\textbf{97.08}&\textbf{99.24}&\textbf{8.93}&\textbf{97.90}&\textbf{97.31}&\textbf{97.72}&\textbf{6.14}&\textbf{99.25}&\textbf{98.45}&\textbf{99.67}&\textbf{3.20} & \\

    \addlinespace

    %   \rowcolor{lightgray}
    
    \bottomrule

\end{tabular*}
\end{table*}

\begin{table}[ht]

%\footnotesize
\small
\caption{\small Ablation study for classification. }
\centering
\begin{tabular}{@{\extracolsep{\fill}}ccccc}
\toprule   
{} &{}&\multicolumn{1}{c}{Accuracy } &&  {Average Accuracy}  \\

 \cmidrule{2-4} 
\centering 
   & PER$_1$  & PER$_2$ & PER$_3$\\ 
\midrule
MP &59.14 &44.33 & 99.42&  67.01\\
 MP+PLs   & \textbf{94.82}& \textbf{99.70}&\textbf{100} & \textbf{98.07}   \\

\bottomrule
\end{tabular}
\label{classification-res}
\end{table}

\section{Experiments and Results}
Our experiments are carried out using an NVIDIA GeForce RTX 3070 GPU, an Intel Core i7-11800H CPU, and a 32GB DDR4 RAM module. 
\subsection{Dataset and Evaluation}
In this study, we constructed our own face dataset using Infineon's BGT60TR13C $\SI{60}{\giga\hertz}$ FMCW radar sensor. The data collection process spanned a six-month period. Each recording was taken at a distance of 25 cm from the radar sensor, capturing a face for a duration of 2 minutes. Notably, none of these participants wore accessories on their faces during the recordings. The recordings took place on different days and at various times of the day, including morning, noon, and afternoon. Moreover, different room backgrounds were incorporated to introduce environmental diversity into the dataset. It consists of two types of samples: ID and OOD. ID samples are from three male participants, who are the first three authors of this study. OODs are from 13 people, including ten males and three females. In total, we acquired 190126 ID frames (balanced) and 15818 OOD frames. For the ID data, 171118 frames are allocated for training and 19008 frames for testing. The dataset will be available here\footnote{\href{https://syncandshare.lrz.de/getlink/fiQoBoPTK8npvXzYdTfr6a/}{https://syncandshare.lrz.de/getlink/fiQoBoPTK8npvXzYdTfr6a/}}. Written consent is available from those involved in the research.

In OOD detection, we employ four standard metrics. \textbf{AUROC} quantifies the area under the receiver operating characteristic (ROC) curve. \textbf{AUPR\textsubscript{IN/OUT}} corresponds to the area under the precision-recall curve, specifically considering ID/OOD samples as positives. \textbf{FPR95}  denotes the false positive rate (FPR) at a true positive rate (TPR) of 95\%. In the context of human face classification, \textbf{accuracy} serves as the evaluation metric. Additionally, we provide \textbf{Test Time}, signifying the inference time in seconds necessary to assess all test samples.

To compare the performance of FOOD, we train the ResNet34\cite{resnet} architecture in a multi-class classification manner. The strong ResNet34\cite{resnet} pipeline provides a slightly better human face classification of 99.10\% than FOOD. However, it does not have the capability to detect the OODs. To compare the OOD detection capability of FOOD to SOTA methods, we use the same pre-trained ResNet34 \cite{resnet} model above. We apply eight different and powerful OOD detectors to the model and compare their results with FOOD. As seen in Table \ref{ood-results}, FOOD outperforms the SOTA methods in terms of widely used OOD detection metrics. 
\subsection{Ablation}
\begin{figure}[ht]

    \centering

    \includegraphics[width=0.7\columnwidth]{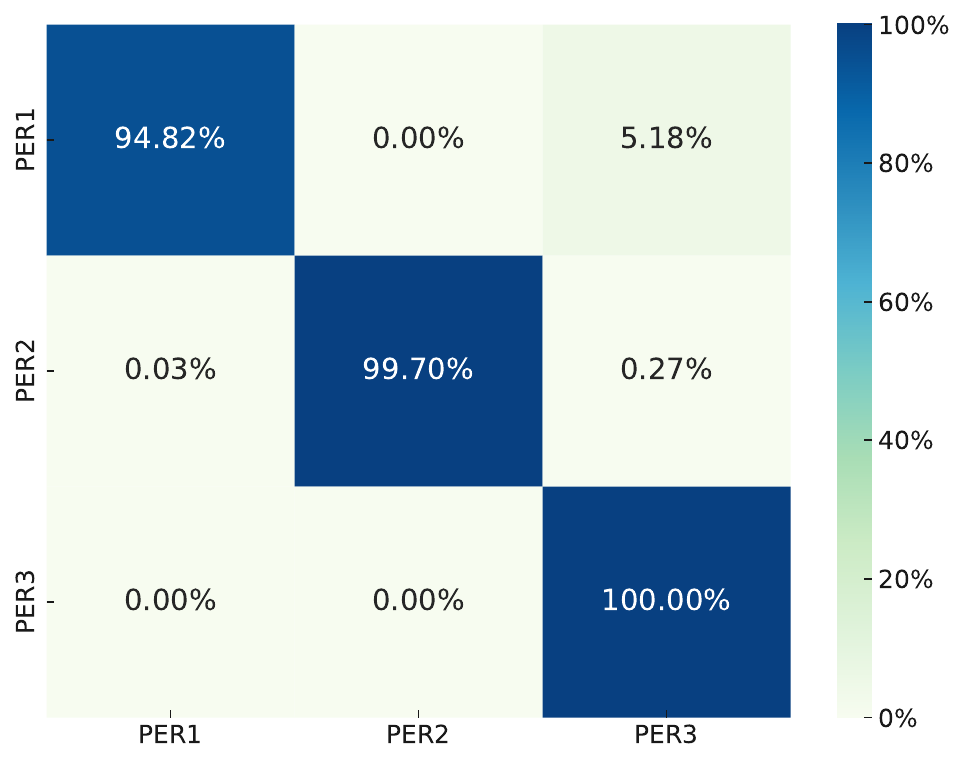}
    \label{fig:conf_matrix_FOOD}
     \caption{\small Confusion matrix to demonstrate the classification performance of FOOD.}
     \label{fig:conf_matrices}
\end{figure}

We also perform ablation studies. Table \ref{classification-res} reflects the impact of using \textbf{MP} (MP$_1$,MP$_2$, and MP$_3$) together with \textbf{PLs} on human face classification accuracy. Figure \ref{fig:conf_matrices} provides a confusion matrix. Additionally, Table \ref{ablation} demonstrates the effects of \textbf{CL} and \textbf{PLs} when they are used together for OOD detection. Based on the application and the level of OOD detection necessity, the \textbf{CL} can be used alone for immediate detection of the OODs because it also provides highly acceptable results. However, \textbf{PLs} are located just before the final layer of the \textbf{MP}'s decoders and include high-level information of their corresponding ID class. Since they are specialized only in their ID class, they greatly impact accurate classification and OOD detection.

\section{Conclusion}
\vspace{-0.2cm}
This paper presents an innovative framework for short-range FMCW radar-based face authentication and OOD detection. The proposed pipeline accurately classifies ID samples and effectively prevents inaccurate predictions for OOD samples. The reconstruction-based architecture, featuring a main convolutional one-encoder multi-decoder and intermediate linear one-encoder one-decoder components, contributes to a highly accurate human face classifier and a robust OOD detector. Our network achieves 98.07\% average classification accuracy for ID human faces and 98.50\% average AUROC as an OOD detector, surpassing previous SOTA approaches. Despite initially using three ID classes, the flexible structure of FOOD allows easy expansion to classify as many ID human faces as needed.

\vfill\pagebreak

\bibliographystyle{ieeetr}

%\bibliographystyle{IEEEbib}
%\bibliography{strings,refs}

\newpage
\footnotesize
\begin{thebibliography}{10}

\bibitem{human_presence7}
Prateek Nallabolu, Li~Zhang, Hong Hong, and Changzhi Li,
\newblock ``Human presence sensing and gesture recognition for smart home applications with moving and stationary clutter suppression using a 60-ghz digital beamforming fmcw radar,''
\newblock {\em IEEE Access}, vol. 9, pp. 72857--72866, 2021.

\bibitem{human_activity3}
Thomas Stadelmayer, Markus Stadelmayer, Avik Santra, Robert Weigel, and Fabian Lurz,
\newblock ``Human activity classification using mm-wave fmcw radar by improved representation learning,''
\newblock in {\em Proceedings of the 4th ACM Workshop on Millimeter-Wave Networks and Sensing Systems}, New York, NY, USA, 2020, mmNets'20, Association for Computing Machinery.

\bibitem{people_counting}
Cem~Yusuf Aydogdu, Souvik Hazra, Avik Santra, and Robert Weigel,
\newblock ``Multi-modal cross learning for improved people counting using short-range fmcw radar,''
\newblock in {\em 2020 IEEE International Radar Conference (RADAR)}, 2020, pp. 250--255.

\bibitem{gesture_recog}
Souvik Hazra and Avik Santra,
\newblock ``Robust gesture recognition using millimetric-wave radar system,''
\newblock {\em IEEE Sensors Letters}, vol. 2, no. 4, pp. 1--4, 2018.

\bibitem{heartbeat_est}
Muhammad Arsalan, Avik Santra, and Christoph Will,
\newblock ``Improved contactless heartbeat estimation in fmcw radar via kalman filter tracking,''
\newblock {\em IEEE Sensors Letters}, vol. 4, no. 5, pp. 1--4, 2020.

\bibitem{b9}
Qing Yu and Kiyoharu Aizawa,
\newblock ``Unsupervised out-of-distribution detection by maximum classifier discrepancy,''
\newblock in {\em IEEE/CVF International Conference on Computer Vision (ICCV)}, 2019.

\bibitem{b5}
Chandramouli~Shama Sastry and Sageev Oore,
\newblock ``Detecting out-of-distribution examples with {G}ram matrices,''
\newblock in {\em International Conference on Machine Learning (ICML)}, 2020.

\bibitem{b27}
Jingkang Yang, Haoqi Wang, Litong Feng, Xiaopeng Yan, Huabin Zheng, Wayne Zhang, and Ziwei Liu,
\newblock ``Semantically coherent out-of-distribution detection,''
\newblock in {\em IEEE International Conference on Computer Vision (ICCV)}, 2021.

\bibitem{b25}
Haoqi Wang, Zhizhong Li, Litong Feng, and Wayne Zhang,
\newblock ``Vim: Out-of-distribution with virtual-logit matching,''
\newblock {\em arXiv}, 2022.

\bibitem{dnn-based}
Hae-Seung Lim, Jaehoon Jung, Jae-Eun Lee, Hyung-Min Park, and Seongwook Lee,
\newblock ``Dnn-based human face classification using 61 ghz fmcw radar sensor,''
\newblock {\em IEEE Sensors Journal}, vol. 20, no. 20, pp. 12217--12224, 2020.

\bibitem{cnn-based}
J.~Kim, J.‐E Lee, H.‐S Lim, and S.~Lee,
\newblock ``Face identification using millimetre-wave radar sensor data,''
\newblock {\em Electronics Letters}, vol. 56, 08 2020.

\bibitem{32by32}
Eran Hof, Amichai Sanderovich, Mohammad Salama, and Evyatar Hemo,
\newblock ``Face verification using mmwave radar sensor,''
\newblock in {\em 2020 International Conference on Artificial Intelligence in Information and Communication (ICAIIC)}, 2020, pp. 320--324.

\bibitem{imp32by32}
Muralidhar~Reddy Challa, Abhinav Kumar, and Linga~Reddy Cenkeramaddi,
\newblock ``Face recognition using mmwave radar imaging,''
\newblock in {\em 2021 IEEE International Symposium on Smart Electronic Systems (iSES)}, 2021, pp. 319--322.

\bibitem{one-shot}
Ha-Anh Pho, Seongwook Lee, Vo-Nguyen Tuyet-Doan, and Yong-Hwa Kim,
\newblock ``Radar-based face recognition: One-shot learning approach,''
\newblock {\em IEEE Sensors Journal}, vol. 21, no. 5, pp. 6335--6341, 2021.

\bibitem{point-cloud-face}
Youxuan Zhong, Chun Yuan, Yi~Zou, and Heng Yao,
\newblock ``Face recognition based on point cloud data captured by low-cost mmwave radar sensors,''
\newblock in {\em 2023 IEEE 13th Annual Computing and Communication Workshop and Conference (CCWC)}, 2023, pp. 0074--0083.

\bibitem{b1}
Dan Hendrycks and Kevin Gimpel,
\newblock ``A baseline for detecting misclassified and out-of-distribution examples in neural networks,''
\newblock in {\em International Conference on Learning Representations (ICLR)}, 2017.

\bibitem{b2}
Shiyu Liang, Yixuan Li, and R.~Srikant,
\newblock ``Enhancing the reliability of out-of-distribution image detection in neural networks,''
\newblock in {\em International Conference on Learning Representations (ICLR)}, 2018.

\bibitem{b3}
Balaji Lakshminarayanan, Alexander Pritzel, and Charles Blundell,
\newblock ``Simple and scalable predictive uncertainty estimation using deep ensembles,''
\newblock in {\em Advances in Neural Information Processing Systems}. 2017, vol.~30, Curran Associates, Inc.

\bibitem{b30}
Y.~C. {Hsu}, Y.~{Shen}, H.~{Jin}, and Z.~{Kira},
\newblock ``Generalized odin: Detecting out-of-distribution image without learning from out-of-distribution data,''
\newblock in {\em IEEE/CVF Conference on Computer Vision and Pattern Recognition (CVPR)}, 2020.

\bibitem{b4}
Kimin Lee, Kibok Lee, Honglak Lee, and Jinwoo Shin,
\newblock ``A simple unified framework for detecting out-of-distribution samples and adversarial attacks,''
\newblock in {\em International Conference on Neural Information Processing Systems (NeurIPS)}, 2018.

\bibitem{b6}
Haiwen Huang, Zhihan Li, Lulu Wang, Sishuo Chen, Bin Dong, and Xinyu Zhou,
\newblock ``Feature space singularity for out-of-distribution detection,''
\newblock in {\em Proceedings of the Workshop on Artificial Intelligence Safety 2021 (SafeAI 2021)}, 2021.

\bibitem{b31}
Yiyou Sun, Yifei Ming, Xiaojin Zhu, and Yixuan Li,
\newblock ``Out-of-distribution detection with deep nearest neighbors,''
\newblock in {\em International Conference on Machine Learning (ICML)}, 2022.

\bibitem{b7}
Weitang Liu, Xiaoyun Wang, John Owens, and Yixuan Li,
\newblock ``Energy-based out-of-distribution detection,''
\newblock in {\em Advances in Neural Information Processing Systems (NeurIPS)}, 2020.

\bibitem{b28}
Yiyou Sun, Chuan Guo, and Yixuan Li,
\newblock ``React: Out-of-distribution detection with rectified activations,''
\newblock in {\em Advances in Neural Information Processing Systems (NeurIPS)}, 2021.

\bibitem{b14}
Rui Huang, Andrew Geng, and Yixuan Li,
\newblock ``On the importance of gradients for detecting distributional shifts in the wild,''
\newblock in {\em Advances in Neural Information Processing Systems (NeurIPS)}, 2021.

\bibitem{hendrycks2022scaling}
Dan Hendrycks, Steven Basart, Mantas Mazeika, Andy Zou, Joe Kwon, Mohammadreza Mostajabi, Jacob Steinhardt, and Dawn Song,
\newblock ``Scaling out-of-distribution detection for real-world settings,''
\newblock {\em International Conference on Machine Learning (ICML)}, 2022.

\bibitem{b8}
Dan Hendrycks, Mantas Mazeika, and Thomas Dietterich,
\newblock ``Deep anomaly detection with outlier exposure,''
\newblock in {\em International Conference on Learning Representations (ICLR)}, 2019.

\bibitem{b10}
Aristotelis-Angelos Papadopoulos, Mohammad~Reza Rajati, Nazim Shaikh, and Jiamian Wang,
\newblock ``Outlier exposure with confidence control for out-of-distribution detection,''
\newblock {\em Neurocomputing}, vol. 441, pp. 138--150, 2021.

\bibitem{RB-OOD}
Sabri~Mustafa Kahya, Muhammet~Sami Yavuz, and Eckehard Steinbach,
\newblock ``Reconstruction-based out-of-distribution detection for short-range fmcw radar,''
\newblock in {\em 2023 31st European Signal Processing Conference (EUSIPCO)}, 2023, pp. 1350--1354.

\bibitem{MCROOD}
Sabri~Mustafa Kahya, Muhammet Sami~Yavuz, and Eckehard Steinbach,
\newblock ``Mcrood: Multi-class radar out-of-distribution detection,''
\newblock in {\em ICASSP 2023 - 2023 IEEE International Conference on Acoustics, Speech and Signal Processing (ICASSP)}, 2023, pp. 1--5.

\bibitem{kahya2023hood}
Sabri~Mustafa Kahya, Muhammet~Sami Yavuz, and Eckehard Steinbach,
\newblock ``Hood: Real-time robust human presence and out-of-distribution detection with low-cost fmcw radar,''
\newblock {\em arXiv}, 2023.

\bibitem{kahya2023harood}
Sabri~Mustafa Kahya, Muhammet Sami~Yavuz, and Eckehard Steinbach,
\newblock ``Harood: Human activity classification and out-of-distribution detection with short-range fmcw radar,''
\newblock in {\em ICASSP 2024 - 2024 IEEE International Conference on Acoustics, Speech and Signal Processing (ICASSP)}, 2024, pp. 6950--6954.

\bibitem{kingma2014adam}
Diederik~P. Kingma and Jimmy Ba,
\newblock ``Adam: A method for stochastic optimization,''
\newblock {\em arXiv preprint arXiv:1412.6980}, 2014.

\bibitem{resnet}
Kaiming He, Xiangyu Zhang, Shaoqing Ren, and Jian Sun,
\newblock ``Deep residual learning for image recognition,''
\newblock in {\em IEEE Conference on Computer Vision and Pattern Recognition (CVPR)}, 2016.

\end{thebibliography}

\end{document}